\documentclass[journal]{IEEEtranTIE}
\usepackage{amsmath,amsfonts}
\usepackage{bm}
\usepackage{bbm}
\usepackage{amssymb}
\usepackage{mathtools}
\usepackage{mathrsfs}
\usepackage{graphicx}
\usepackage{algorithmic}
\usepackage{array}
\usepackage[caption=false,font=normalsize,labelfont=sf,textfont=sf]{subfig}
\usepackage{textcomp}
\usepackage{stfloats}
\usepackage{url}
\usepackage{verbatim}
\usepackage{graphicx}
\usepackage{siunitx}
\usepackage{booktabs}
\usepackage{tikz}
\usepackage{nicematrix}

\DeclareSIUnit{\hertz}{Hz}
\def\BibTeX{{\rm B\kern-.05em{\sc i\kern-.025em b}\kern-.08em
    T\kern-.1667em\lower.7ex\hbox{E}\kern-.125emX}}
\usepackage{balance}
\usepackage{xcolor}
\DeclareMathOperator*{\mR}{\mathbb{R}}
\DeclareMathOperator*{\mE}{\mathbb{E}}
\DeclareMathOperator*{\SO}{SO}

\usepackage{cite}
\usepackage{picinpar}
\usepackage{amsmath}
\usepackage{url}
\usepackage{flushend}
\usepackage[utf8]{inputenc}
\usepackage{colortbl}
\usepackage{soul}
\usepackage{multirow}
\usepackage{pifont}
\usepackage{color}
\usepackage{alltt}
\usepackage[hidelinks]{hyperref}
\usepackage{enumerate}
\usepackage{siunitx}
\usepackage{breakurl}
\usepackage{epstopdf}
\usepackage{pbox}

\usepackage[percent]{overpic} 
\usepackage{xcolor}

\definecolor{lightblue}{RGB}{70,130,255}

\usepackage{placeins}

\usepackage{afterpage}

\begin{document}
\title{Terrain-Awared LiDAR-Inertial Odometry for Legged-Wheel Robots Based on Radial Basis Function Approximation}

\author{
	\vskip 1em
	
	Yizhe Liu,
	Han Zhang

	 \thanks{
	
	 	Manuscript received Month xx, 2xxx; revised Month xx, xxxx; accepted Month x, xxxx.
		
        Both authors are with School of Automation and Intelligent Sensing, Institute of Medical Robotics, Shanghai Jiao Tong University, and Key Laboratory of System Control and Information Processing, Ministry of Education, Shanghai 200240, China. Corresponding author: Han Zhang (e-mail: zhanghan\_tc@sjtu.edu.cn).
	 }
}

\maketitle
	
\begin{abstract}
An accurate odometry is essential for legged-wheel robots operating in unstructured terrains such as bumpy roads and staircases. Existing methods often suffer from pose drift due to their ignorance of terrain geometry. 
We propose a terrain-awared LiDAR-Inertial  odometry (LIO) framework that approximates the terrain using Radial Basis Functions (RBF) whose centers are adaptively selected and weights are recursively updated.
The resulting smooth terrain manifold enables ``soft constraints" that regularize the odometry optimization and mitigates the $z$-axis pose drift under abrupt elevation changes during robot's maneuver. To ensure the LIO's real-time performance, we further evaluate the RBF-related terms and calculate the inverse of the sparse kernel matrix with GPU parallelization.
Experiments on unstructured terrains demonstrate that our method achieves higher localization accuracy than the state-of-the-art baselines, especially in the scenarios that have continuous height changes or sparse features when abrupt height changes occur.
\end{abstract}


\markboth{IEEE TRANSACTIONS ON INDUSTRIAL ELECTRONICS}%
{}

\definecolor{limegreen}{rgb}{0.2, 0.8, 0.2}
\definecolor{forestgreen}{rgb}{0.13, 0.55, 0.13}
\definecolor{greenhtml}{rgb}{0.0, 0.5, 0.0}

\section{Introduction}

\IEEEPARstart{L}{egged-wheel} robots combine the speed advantage of wheeled robots with the terrain adaptability advantage of legged robots. Thus, they are well-suited for traversing complex and uneven environments such as bumpy roads, staircases, etc.
However, the uneven surface in these environments will cause impulsive velocity variations during the robot's maneuver. 
Such jolts will pose significant challenges to the odometry systems, resulting in pose estimation drifts, particularly in the $z$-axis. 
Therefore, the terrain geometry needs to be carefully taken into account when designing the odometry module.
Earlier studies usually model the terrain as discrete 2.5-D elevation maps. In particular, they capture the overall elevations in a mesh, but ignore the small elevation changes within each grid. Indeed, setting a smaller grid resolution can depict more details
but 2.5-D elevation maps are not spatially continuous and hence not spatially differentiable. Thus, it cannot be used directly in the optimization process of an odometry framework and we need to consider a different terrain representation.

To this end, we use Radial Basis Functions (RBF) to approximate the terrain.
Compared to 2.5-D elevation maps, it provides a smooth representation for the terrain based on LiDAR point clouds.
Based on the moment conditions, we further adopt a recursive ridge regression in Kalman filter style to update the RBF weights as new data arrives. 
Hence as the robot moves, the new observations will be continuously fused into the terrain approximation.

Now that the RBF-based terrain approximation forms a smooth manifold whose gradient can be explicitly stated, we further introduce  ``soft constraints" based on the approximated terrain manifold into the optimization of the LiDAR-Inertial odometry (LIO) framework.
Consequently, the ``soft constraints" augments the scan matching objective function with a dedicated cost term to build a more robust odometry.
In particular, when abrupt height changes occur, the manifold constraint can anchor the robot's vertical pose to the approximated terrain surface and reduce $z$-axis pose drift. 
To further improve the real-time performance of our LIO framework, both the RBF-related term evaluation and the sparse kernel matrix inversion are implemented in CUDA-based GPU kernels. 

\begin{figure}[!htpb]
  \centering
  \includegraphics[width=0.7\linewidth]{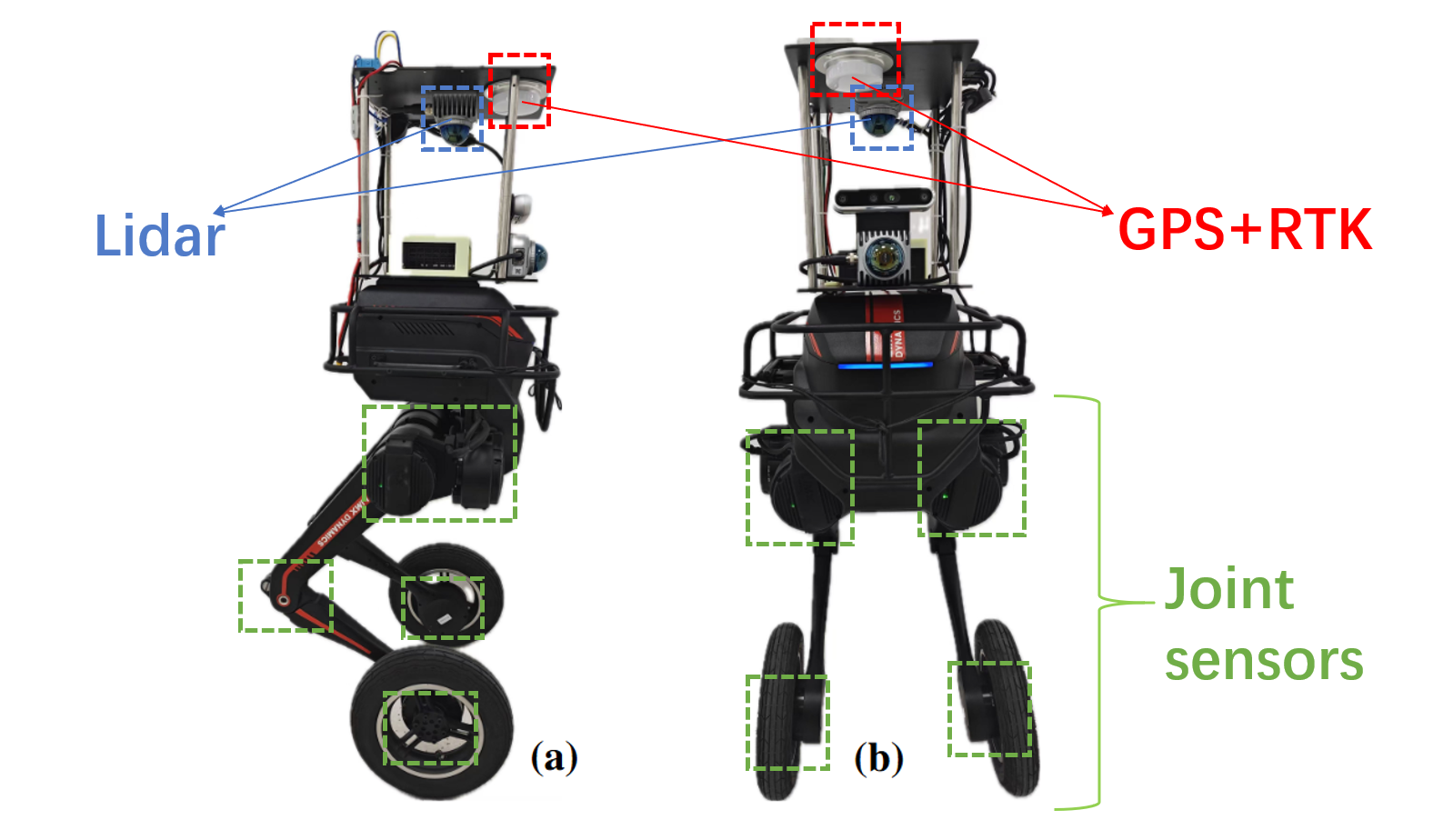}
  \caption{Robot platform used in our experiments. The system mounts two LiDARs; however, only the top, inverted Mid360 LiDAR (blue) is used in all experiments (the lower LiDAR is not used). The platform is also equipped with a GPS+RTK module (red) and joint sensors (green) for kinematic state estimation. (a) side view, (b) front view.}
  \label{fig:robot-setting}
\end{figure}

To summarize, the main contribution of this work is three-fold:
\begin{itemize}
    \item We approximate the uneven terrains with RBF functions. This gives a smooth terrain approximation whose gradient can be explicitly stated.
    Moreover, based on the moment conditions, a recursive ridge regression is used to update the RBF weights. Furthermore, we use an adaptive RBF center selection strategy and GPU parallelization to accelerate the computation. 
    \item We construct ``soft constraints" based on the approximated terrain and introduce them into the LIO optimization to suppress the vertical drift. In addition, we implement the RBF-related terms evaluation using GPU parallelization within the scan matching optimization module.
    \item 
    Experiments have been conducted to test the performance of our proposed LIO framework. Consequently, we outperform ROLO-SLAM \cite{roloslam} and FAST-LIO2 \cite{fastlio2} on uneven terrains, particularly in the scenarios that have continuous height changes or sparse features when abrupt height changes occur.
    In addition, we also release the dataset \footnote{Available at https://zhanghan-tc.github.io/legged\_wheel\_dataset.} collected in this study, which provides a resource for evaluating the SLAM dedicated for legged-wheel robots.
\end{itemize}

\section{Related Work}
LIO has been widely studied in recent years. LOAM \cite{loam} achieves an accurate real-time odometry and mapping with LiDAR-only input. Later works, such as LIO-SAM \cite{liosam} and Fast-LIO2 \cite{fastlio2, fastlio}, extend this line of research by tightly coupling the LiDAR and the IMU measurements, which improves both the robustness and the accuracy in various environments. However, they often suffer from pose drifts when operating in height-varying or unstructured terrains. 
Leg-KILO \cite{leg-kilo} introduces leg kinematic and extracts LiDAR point around the foot contact point to construct contact plane constraints to improve the accuracy. 
In particular, it assumes that the terrain that is around the feet's contact points is flat within a radius of 10 cm.
However, such contact planes may not be sufficient to approximate the uneven terrain well and degrade the LIO's accuracy.
ROLO-SLAM \cite{roloslam} estimates robot's rotation first, and then estimate its translation under the rotation constraint. 
However, for legged–wheel robots that operates on uneven terrains, its rotation and translation are strongly interdependent due to its frequent body tilts. Hence such a decoupled strategy often leads to severe drift and degraded robustness.

These existing works give us a strong motivation to approximate the terrain geometry and utilize the geometry in LIO design to improve its accuracy. 
To better capture the geometry of uneven environments, elevation mapping has been widely adopted. Early works on robot-centric elevation maps \cite{elevation_map_2014} and probabilistic terrain representations \cite{elevation_map_2018} improve the robot's terrain awareness by incorporating the elevation uncertainty. Later methods exploit GPU acceleration for real-time elevation mapping \cite{elevation_map_gpu} or employ Gaussian Process regression for probabilistic surface estimation \cite{elevation_map_gp}. 
Moreover, utilizing the discrete elevation mapping, \cite{vilen} incorporate foot-ground contacts to constrain state estimation, \cite{lio_gc} leverages the slope from the discrete elevation mapping to construct graph factors. 
However, elevation maps are posed on a mesh that has a fixed resolution. Consequently, the height is not spatially differentiable.
As a result, these constraints and factors in \cite{vilen}\cite{lio_gc} are typically derived from discretized grids or planar patches, which yield piecewise, non-smooth gradient information that is unreliable for optimization.

On the other hand, RBFs offer an alternative for surface reconstruction. In particular, it produces continuous and dense terrain model, and have been successfully applied to 3D shape modeling \cite{rbf2001} and digital terrain interpolation from airborne LiDAR data \cite{rbf_air}. Consequently, RBF-based terrain approximation provides a continuous and differentiable model that can be seamlessly integrated to optimization-based estimation. Nevertheless, as the LiDAR measurements accumulate over time, direct least-squares estimation of the RBF weights will be computationally expensive.  
Recursive approaches provide a systematic way to address this challenge. The connection between Recursive Least Squares (RLS) and the Kalman filter has been well established in optimization theory \cite{OptVec}, enabling incremental updates of the model parameters as new data arrives. This perspective has been adopted in online regression and adaptive filtering, but to the best of our knowledge, its application to terrain modeling within the LiDAR-inertial odometry remains underexplored.  

Building upon these insights, we are motivated to build a terrain-awared LIO framework so that it is robust against height changes during robot's maneuver on uneven and unstructured terrains.
More specifically, we aim to build the approximation model for terrain geometry and use the model in the gradient-based LIO optimization, so that it can mitigate the $z$-axis pose drift when abrupt elevation changes occurs.

\section{The RBF-based Terrain Approximation}\label{sec:RBF_approximation}
In this section, we will introduce the RBF-based terrain approximation method. In particular, we will first introduce the RBF approximation model for the terrain.
Then, we will present the recursive ridge regression used for the weights update. 
Note that, in this section, the robot's base pose is assumed to be known.
\subsection{The RBF Approximation Model}\label{sec: RBF_approximation_model}
The RBF approximation is universal \cite{rbf2001}, i.e., given enough centers, it can approximate any continuous function on a compact set to arbitrary accuracy.  
Consequently, the RBF approximation model gives a dense, smooth representation for the terrain.
Compared to 2.5-D elevation maps, the RBF approximation model predicts the height at any arbitrary point within the compact Region of Interest (ROI) $\mathcal{X}$.
Moreover, its spatial gradient can be calculated explicitly at any point, which is essential for solving the optimization problem for the LIO.

To this end, we let the RBF approximation for the terrain be
\begin{equation}
    f(\mathbf{x};\mathbf{w}) = \sum_{i=1}^{N} w_i \, \phi(\lVert \mathbf{x} - \mathbf{c}_i \rVert_2),
    \label{RBF_fml}
\end{equation}
where
\begin{itemize}
    \item \( \mathbf{x} =[p_x,p_y]^\top\in \mathbb{R}^2 \) is a 2D query point in the horizontal plane, and \( f(\mathbf{x};\mathbf{w}) \) denotes the predicted terrain height at the query point;
    \item \( \mathbf{c}_{1:N}:=\{ \mathbf{c}_i = (x_i^c, y_i^c)\in\mathcal{X}\}_{i=1}^{N} \subset \mathbb{R}^2 \) are the RBF center points, which will be selected adaptively based on the distribution of the terrain point cloud;
    \item \(\mathbf{w}:= \{w_i \in \mathbb{R}\}_{i=1}^N \) are the weights of the RBFs;
    \item \( \phi(\cdot) \) is the RBF. In this work, we use the Gaussian kernel, namely,
    \begin{align}
        \phi(r_i) = \exp\left(-\frac{(r_i)^2}{2\sigma^2}\right):=\kappa_{\sigma}(\mathbf{x};\mathbf{c}_j),\label{eq:RBF_def}
    \end{align}
    where \( r_i = \lVert \mathbf{x} - \mathbf{c}_i \rVert_2 \) is the Euclidean distance between the query point $\mathbf{x}$ and center $\mathbf{c}_i$, and \( \sigma\in \mR\) is a bandwidth parameter that determines the spatial influence of the kernel.
\end{itemize}
Notably, the model \eqref{RBF_fml} predicts the terrain height as a continuous function of 2-D spatial coordinates.
Also foreseeably, the RBF center points will affect the approximation accuracy to great extent.
In particular, to achieve a more accurate approximation, we need to add more centers. But given the fact that there are only a finite number of LiDAR points measured from the terrain, an excessive number of center points will not only increase the computation cost but also lead to numerical instability when estimating the weights.
Moreover, since the exponential function in the Gaussian kernel decays very fast as $r_i$ increases, most entries of $\mathbf{m}(\mathbf{x}_k^i)$ in Sec. \ref{sec_LS} to come would be almost zero. Consequently, this would lead to an sparse and ill-conditioned $\mathbf{H}_k$ and affect the approximation accuracy. Hence the amount and the placement of the centers need to be carefully considered to give an accurate approximation of the terrain.

To this end, we use an adaptive strategy to select the RBF centers. 
First, we build a k-d tree from the terrain point cloud for efficient neighborhood queries.
Next, we construct the candidate centers on a mesh with a specified resolution in the compact ROI $\mathcal{X}$ of the $x$-$o$-$y$ plane. 
For each candidate center, the k-d tree counts the number of LiDAR points around the candidate center within a radius $r_{kd}$. If the number exceeds a pre-defined threshold, then the candidate center will be accepted. 
Such a strategy balances the needs for RBF centers' number suppression with the terrain approximation accuracy.

\subsection{The RBF Weight Estimate by Ridge Regression} \label{sec_LS}
To use \eqref{RBF_fml} to approximate the terrain, we need to estimate the RBF weights. 
When there is no prior information available for the terrain (mostly for the first frame), we estimate the RBF weights by solving a ridge regression.
More specifically, since in this section, the robot's base pose is assumed to be known, 
let $\{ \mathbf{p}^i_k \in \mathbb{R}^3 \}_{i=1}^{M_k}$, $(i=1,\ldots,M_k)$ be the LiDAR point cloud data at the $k$-th time step expressed in the world frame, where each point \( \mathbf{p}^i_k := [\mathbf{x}_k^{i\top}, p_{z,k}^i]^\top\), and $\mathbf{x}_k^i:=[p_{x,k}^i, p_{y,k}^i]^\top\in\mR^2$.
In addition, we assume that the points in the observed LiDAR point cloud $\mathbf{p}_k^i$, $\forall i,k$ expressed in the world frame, are all i.i.d. samples of the stochastic vector $\mathbf{p}$. More specifically, it is assumed that 
\begin{align*}
    \mathbf{p}:=\begin{bmatrix}
        \mathbf{x}\\p_z
    \end{bmatrix}=\bar{\mathbf{p}}+\bm\epsilon=\begin{bmatrix}\bar{\mathbf{x}}\\\bar{p}_{z}\end{bmatrix}+\begin{bmatrix}
        \epsilon_{\mathbf{x}}\\
        \epsilon_z
    \end{bmatrix}, 
\end{align*}
where $\bar{\mathbf{p}}$ is a stochastic vector of 3D-coordinates which is randomly drawn from the terrain surface expressed in the world frame, $\bar{\mathbf{x}}=[p_{x},p_{y}]^\top$, and $\bm\epsilon\sim\mathcal{N}(0,\sigma_\epsilon^2I)$ is the point cloud measurement noise and it is independent of $\bar{\mathbf{p}}$.
Further, suppose \eqref{RBF_fml} approximates the terrain surface well on $\mathcal{X}$, i.e., it holds that
\begin{align}\label{eq:fit_height}
    p_z = \underbrace{\begin{bmatrix}
        \kappa_\sigma(\bar{\mathbf{x}},\mathbf{c_1})&\cdots &\kappa_\sigma(\bar{\mathbf{x}},\mathbf{c}_N)
    \end{bmatrix}}_{\bm{\Phi}(\mathbf{\bar{x}})^\top}
    \underbrace{\begin{bmatrix}
        w_1\\\vdots\\w_N
    \end{bmatrix}}_{\mathbf{w}}+\epsilon_z,
\end{align}
where the centers $\mathbf{c}_j\in\mathcal{X},i=1,\ldots,N$ and $\mathbf{w}$ is the RBF weights.

Next, recall that our goal is to fit the terrain surface by estimating the RBF weights $\mathbf{w}$. 
Nevertheless, we do not have access to the i.i.d. sample of $\bar{\mathbf{x}}$, but only have the i.i.d samples $\mathbf{x}^i_k$ of $\mathbf{x}$ instead. 
To this end, let $\mathbf{m}(\mathbf{x}):=\mE[\bm\Phi(\mathbf{\bar{x}})|\mathbf{x}]$. Notably, $\epsilon_z$ is independent of $\mathbf{x}$, hence it holds that 
\begin{align*}
    \mE[p_z|\mathbf{x}]=\mE[\bm\Phi(\mathbf{\bar{x}})^\top\mathbf{w}+\epsilon_z|\mathbf{x}]=\mE[\bm\Phi(\mathbf{\bar{x}})^\top|\mathbf{x}]\mathbf{w}=\mathbf{m}(\mathbf{x})^\top\mathbf{w}.
\end{align*}
Therefore, by the tower property of conditional expectation, it follows that 
\begin{equation}
\begin{aligned}
    \mE[\mathbf{m}(\mathbf{x})p_z]&=\mE[\mE[\mathbf{m}(\mathbf{x})p_z|\mathbf{x}]]=\mE[\mathbf{m}(\mathbf{x})\mE[p_z|\mathbf{x}]]\\
    &=\mE[\mathbf{m}(\mathbf{x})\mathbf{m}(\mathbf{x})^\top]\mathbf{w}.
    \label{eq:moment_equality}
\end{aligned}
\end{equation}
Notably, \eqref{eq:moment_equality} forms an moment equality and we can get the estimator at the $k$-th time step
\begin{align}
&\hat{\mathbf{w}}_k=\mE[\mathbf{m}(\mathbf{x})\mathbf{m}(\mathbf{x})^\top]^{-1}\mE[\mathbf{m}(\mathbf{x})p_z]\nonumber\\
&\approx (\underbrace{\sum_{t=1}^k\sum_{i=1}^{M_t} \mathbf{m}(\mathbf{x}_t^i)\mathbf{m}(\mathbf{x}_t^i)^\top}_{\mathbf{H}_k})^{-1}(\underbrace{\sum_{t=1}^k\sum_{i=1}^{M_t}  \mathbf{m}(\mathbf{x}_t^i)p_{z,t}^i}_{\mathbf{b}_k}),\label{eq:RBF_estimator}
\end{align}
and 
\begin{align*}
    \mathbf{m}(\mathbf{x}_t^i)& = \mE[\bm{\Phi}(\bar{\mathbf{x}})|\mathbf{x}_t^i] \\
    & =\frac{\sigma^2}{\tilde{\sigma}^2}\begin{bmatrix}
        \kappa_{\tilde{\sigma}}(\mathbf{x}_t^i;\mathbf{c}_1) &\cdots & \kappa_{\tilde{\sigma}}(\mathbf{x}_t^i;\mathbf{c}_N)
    \end{bmatrix}^\top,
\end{align*}
where $\tilde{\sigma}=\sqrt{\sigma^2+\sigma_\epsilon^2}$, since $\bm\epsilon\sim\mathcal{N}(0,\sigma^2_\epsilon I)$.
Notably, \eqref{eq:RBF_estimator} is the unique minimizer of 
\begin{align}
    \min_{\mathbf{w}} & \sum_{t=1}^k\sum_{i=1}^{M_t}(p_{z,t}^i-\mathbf{m}(\mathbf{x}_t^i)^\top \mathbf{w})^2.\label{eq:mm_corresponding_obj_fun}
\end{align}
Nevertheless, as mentioned in Sec.~\ref{sec: RBF_approximation_model}, $\mathbf{
H}_k$ might be ill-conditioned due to the intrinsics of RBF and greatly affect the approximation accuracy. To address this issue, we add a regularization term to \eqref{eq:mm_corresponding_obj_fun} and do a ridge regression \cite{ridge} instead, i.e.,
\begin{align*}
\min_{\mathbf{w}} & \sum_{t=1}^k\sum_{i=1}^{M_t}(p_{z,t}^i-\mathbf{m}(\mathbf{x}_t^i)^\top \mathbf{w})^2 + \lambda \|\mathbf{w}\|^2,
\end{align*}
and the corresponding estimator is $\hat{\mathbf{w}}^{RG}_k=(\underbrace{\mathbf{H}_k+\lambda I}_{\bm{\mathcal{H}}_k})^{-1}b_k$.
In particular, the ridge penalty $\lambda \|\mathbf{w}\|^2$ mitigates the ill-conditioning of $\mathbf{H}_k$ and numerically stabilizes the least square estimate of \eqref{eq:mm_corresponding_obj_fun}. However, it reduces variance at the expense of introducing bias. 
Nevertheless, such bias is typically modest and does not inflate the residual sum of squares unreasonably, while the reduction in variance can yield estimators with small mean-squared error \cite{ridge}.

\subsection{The Recursive RBF Weight Update}\label{recur_RBF}
Recall the terrain approximation model \eqref{RBF_fml}.
Since the RBF \eqref{eq:RBF_def} decays exponentially to zero as $r_i$ increases, the RBF $\kappa_\sigma(\mathbf{x};\mathbf{c}_i)$ only captures the terrain geometry in the vicinity of $\mathbf{c}_i\in\mathcal{X}$. 
On the other hand, according to \eqref{eq:RBF_estimator}, the updated estimator $\hat{\mathbf{w}}_{k+1}$ from the LiDAR new point cloud measurement at the new time step $k+1$ shall take the form
\begin{subequations}
\begin{align}
\hat{\mathbf{w}}_{k+1}^{RG}
&=(\bm{\mathcal{H}}_{k+1})^{-1}(\mathbf{b}_{k}+\mathfrak{m}_{k+1}\mathbf{p}_{z,k+1}),
\label{eq:weights_update}\\
\bm{\mathcal{H}}_{k+1}&=\bm{\mathcal{H}}_k+\mathfrak{m}_{k+1}\mathfrak{m}_{k+1}^\top\label{eq:H_update}
\end{align}
\end{subequations}
where
and $\mathfrak{m}_{k+1}\in\mR^{N\times M_{k+1}}$, $\mathbf{p}_{z,k+1}\in\mR^{M_{k+1}}$. More specifically,
\begin{subequations}
\begin{align}
    &\mathfrak{m}_{k+1}:=\begin{bmatrix}
        \mathbf{m}(\mathbf{x}_{k+1}^1) & \ldots &\mathbf{m}(\mathbf{x}_{k+1}^{M_{k+1}})
    \end{bmatrix}\nonumber\\&=\frac{\sigma^2}{\tilde\sigma^2}\begin{bmatrix}
        \kappa_\sigma(\mathbf{x}_{k+1}^1,\mathbf{c}_1) &\cdots &\kappa_\sigma(\mathbf{x}_{k+1}^{M_{k+1}},\mathbf{c}_1)\\
        \vdots &\cdots &\vdots\\
        \kappa_\sigma(\mathbf{x}_{k+1}^1,\mathbf{c}_N) &\cdots &\kappa_\sigma(\mathbf{x}_{k+1}^{M_{k+1}},\mathbf{c}_N)
    \end{bmatrix}\label{eq:rbf_kernel}\\
    &\approx\frac{\sigma^2}{\tilde\sigma^2}\begin{bmatrix}
        \mathbf{0} \\\bm\kappa_\sigma(\mathbf{x}_{k+1}^{1:M_{k+1}},\mathbf{c}_{j_{1,k+1}}) \\ \vdots \\ \bm\kappa_\sigma(\mathbf{x}_{k+1}^{M_{k+1}},\mathbf{c}_{j_{2,k+1}}) \\ \mathbf{0}
    \end{bmatrix}=\begin{bmatrix}
        \mathbf{0}\\\tilde{\mathfrak{m}}_{k+1}\\\mathbf{0}
    \end{bmatrix},\label{eq:mfk_m_approx}\\
    &\mathbf{p}_{z,k+1}:=\begin{bmatrix}
        p_{z,k+1}^1 &\cdots &p_{z,k+1}^{M_{k+1}}
    \end{bmatrix}^\top.
\end{align}
\end{subequations}
Notably, the approximation \eqref{eq:mfk_m_approx} lie in the fact that the LiDAR scans can only cover a local portion of the entire ROI $\mathcal{X}$, and $\kappa_\sigma(\mathbf{x},\mathbf{c}_j)$ decays exponentially as the distance between $\mathbf{x}$ and $\mathbf{c}_j$ increases. Hence most rows of matrix $\mathfrak{m}_{k+1}$ are ``zeros" numerically except for some rows whose corresponding centers have point clouds $\mathbf{x}_t^i$ in the vicinity. (And we refer these rows as ``\textit{active rows}".)
Moreover, in view of \eqref{eq:H_update}, $\bm{\mathcal{H}}_k$ should be numerically block diagonal for all $k$ since we add a matrix whose entries are all zeros except for the block on the diagonal at every time step $k$. Furthermore, since $\mathbf{b}_k=\bm{\mathcal{H}}_k^{-1}\hat{\mathbf{w}}_k$ and in view of \eqref{eq:H_update}, we can rewrite \eqref{eq:weights_update} in a Kalman filter style as
\begin{align}
    &\hat{\mathbf{w}}_{k+1}^{RG} = \bm{\mathcal{H}}_{k+1}^{-1}\mathbf{b}_k+\bm{\mathcal{H}}_{k+1}^{-1}\mathfrak{m}_{k+1}\mathbf{p}_{z,k+1}\nonumber\\
    &=\bm{\mathcal{H}}_{k+1}^{-1}\bm{\mathcal{H}}_{k}\hat{\mathbf{w}}_k+\bm{\mathcal{H}}_{k+1}^{-1}\mathfrak{m}_{k+1}\mathbf{p}_{z,k+1}\nonumber\\
    &=\bm{\mathcal{H}}_{k+1}^{-1}(\bm{\mathcal{H}}_{k+1}^{-1}-\mathfrak{m}_{k+1}\mathfrak{m}_{k+1}^\top)\hat{\mathbf{w}}_k+\bm{\mathcal{H}}_{k+1}^{-1}\mathfrak{m}_{k+1}\mathbf{p}_{z,k+1}\nonumber\\
    &=\hat{\mathbf{w}}_k^{RG}+\bm{\mathcal{H}}_{k+1}^{-1}\mathfrak{m}_{k+1}[\mathbf{p}_{z,k+1}-\mathfrak{m}_{k+1}^\top\hat{\mathbf{w}}_k^{RG}].\label{eq:kf_w_update}
\end{align}
Due to the sparse structure of $\mathfrak{m}_{k+1}$ as illustrated in \eqref{eq:mfk_m_approx} and the block diagonal structure of $\bm{\mathcal{H}}_{k+1}$ (and hence $\bm{\mathcal{H}}_{k+1}^{-1}$ is also block diagonal), we can divide $\hat{\mathbf{w}}_{k+1}$ into blocks. In particular, we can rewrite \eqref{eq:kf_w_update} as
\begin{align}
    &\begin{bmatrix}
        \hat{\mathbf{w}}_{1,k+1}^{RG}\\
        \hat{\mathbf{w}}_{2,k+1}^{RG}\\
        \hat{\mathbf{w}}_{3,k+1}^{RG}
    \end{bmatrix}\!=\!
    \begin{bmatrix}
        \hat{\mathbf{w}}_{1,k}^{RG}\\
        \hat{\mathbf{w}}_{2,k}^{RG}\\
        \hat{\mathbf{w}}_{3,k}^{RG}
    \end{bmatrix}
    \!+\!
    \begin{bmatrix}
        \bm{\mathcal{H}}_{1,k+1}^{-1}\\
        &\bm{\mathcal{H}}_{2,k+1}^{-1}\\
        &&\bm{\mathcal{H}}_{3,k+1}^{-1}
    \end{bmatrix}\!\!\!\!
    \begin{bmatrix}
        \mathbf{0}\\\tilde{\mathfrak{m}}_{k+1}\\\mathbf{0}
    \end{bmatrix}\nonumber\\
    &\times[\mathbf{p}_{z,k+1}-\begin{bmatrix}
        \mathbf{0}^\top &\tilde{\mathfrak{m}}_{k+1} &\mathbf{0}^\top
    \end{bmatrix}
    \begin{bmatrix}
        \hat{\mathbf{w}}_{1,k}^{RG}\\
        \hat{\mathbf{w}}_{2,k}^{RG}\\
        \hat{\mathbf{w}}_{3,k}^{RG}
    \end{bmatrix}]\nonumber\\
    &=
    \begin{bmatrix}
        \hat{\mathbf{w}}_{1,k}^{RG}\\
        \hat{\mathbf{w}}_{2,k}^{RG}+\bm{\mathcal{H}}_{2,k+1}^{-1}\tilde{\mathfrak{m}}_{k+1}(\mathbf{p}_{z,k+1}-\tilde{\mathfrak{m}}_{k+1}\hat{\mathbf{w}}_{2,k}^{RG})\\
        \hat{\mathbf{w}}_{3,k}^{RG}
    \end{bmatrix}.
    \label{eq:active_rows_weights_update}
\end{align}

Moreover, by Woodbury Matrix Identity and in view of the sparse structure, it holds that
\begin{align}
&\bm{\mathcal{H}}_{k\!+\!1}^{-1}\!=\!\bm{\mathcal{H}}_{k}^{-1}\!\!\!\!-\!\bm{\mathcal{H}}_{k}^{-1}\mathfrak{m}_{k\!+\!1}(I\!+\!\mathfrak{m}_{k\!+\!1}^\top\bm{\mathcal{H}}_k^{-1}\mathfrak{m}_{k\!+\!1})^{-1}\mathfrak{m}_{k\!+\!1}^\top
\bm{\mathcal{H}}_{k}^{-1}\nonumber\\
&\!=\!\begin{bmatrix}
    \bm{\mathcal{H}}_{1,k}^{-1}\\
    &\bm{\mathcal{H}}_{2,k+1}^{-1}\\
    &&\bm{\mathcal{H}}_{3,k}^{-1}
\end{bmatrix},
    \label{eq:H_inverse_update}
\end{align}
where $\bm{\mathcal{H}}_{2,k+1}^{-1}:=\bm{\mathcal{H}}_{2,k+1}^{-1}-\bm{\mathcal{H}}_{2,k+1}^{-1}\tilde{\mathfrak{m}}_{k+1}(I+\tilde{\mathfrak{m}}_{k+1}^\top\bm{\mathcal{H}}_{2,k+1}^{-1}\tilde{\mathfrak{m}}_{k+1})^{-1}\tilde{\mathfrak{m}}_{k+1}^\top\bm{\mathcal{H}}_{2,k+1}^{-1}$.
Notably, the recursion \eqref{eq:active_rows_weights_update}, \eqref{eq:H_inverse_update} implies that the only weights update that happens is on those weights that on the ``\textit{active rows}". This means that we only need to run a Kalman filter on the ``\textit{active rows}" to update the weights and hence cut down the computational load by using the sparse structure.

\section{The LIO with the Manifold Constraint}
In previous section, we have obtained an approximation of the terrain. Next, we will use such an approximation as the ``soft constraints" in the LIO system to improve the localization accuracy for a legged-wheel robot that maneuvers on the terrain. Such constraints can introduce additional geometric regularization when estimating the robot base pose, particularly in the vertical direction.

\begin{figure}[!htpb]
    \centering
    \includegraphics[width=0.5\linewidth]{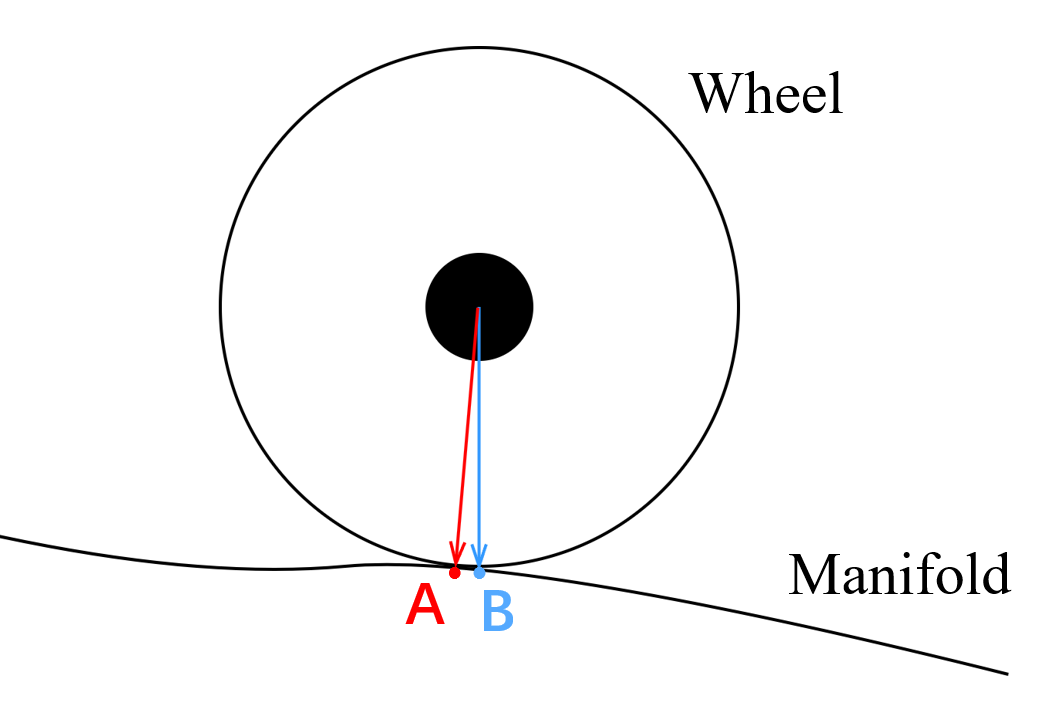}
    \caption{Illustration of wheel–manifold contact. 
    Point A (red) denotes the actual contact point, while Point B (blue) denotes the approximated contact point. 
    This approximation simplifies computation.}
    \label{fig:wheel}
\end{figure}
\subsection{The Leg Kinematics}
To exploit the manifold constraints, we have to estimate the position of the wheel contact point first.
To this end, let \( \mathbf{q}_k \) denote the joint configuration at the time step $k$. We can easily calculate the left/right wheel centers by the forward kinematics of the left leg $\mathbf{h}_\text{L}(\mathbf{q})$ and the right leg $\mathbf{h}_\text{R}(\mathbf{q})$, and hence the coordinate of left wheel center $\bm\xi_\text{L}$ expressed in the world frame shall take the form
\begin{align}\label{eq:leg_fk}
    \bm\xi_\text{L}:=\begin{bmatrix}
        \xi_{x,\text{L}}\\\xi_{y,\text{L}} \\ \xi_{z,\text{L}}
    \end{bmatrix}
    ={}^{\text{B}}\mathbf{R}_\text{O} \mathbf{h}_\text{L}(\mathbf{q})+{}^{\text{B}}\mathbf{t}_{\text{O}},
\end{align}
where ${}^{\text{B}}\mathbf{R}_\text{O}\in\SO(3)$ and ${}^{B}\mathbf{t}_{\text{O}}$ denote the rotational matrix and translational vector that transform the world frame to the base frame. And the right wheel center can be expressed in a similar manner.
Moreover, let $r$ be the wheel radius. We assume that the contact point is approximately right under the wheel center (See Fig. \ref{fig:wheel}). 
Of course, such an approximation is mainly motivated from balancing the computation cost and the accuracy, and the approximation error may increase as the wheel radius $r$ increases. The approximation error would also increase if the robot is standing on very steep slopes. Moreover, the case of multiple contact points are not considered. Nevertheless, as we shall see in the experiment in Sec.~\ref{sec:experiment}, such an approximation is enough to provide an accurate estimate for the base pose. 

\subsection{IMU pre-integration}\label{sec:imu-preint}
Due to the relative low frame rate of LiDAR scans, direct scan matching without a good initial guess is prone to convergence issues, especially under abrupt motions. 
To mitigate this issue, we use an IMU pre-integration module to integrate the high-rate inertial measurements between two consecutive LiDAR frames.
To this end, let the states $\bm\eta_k$ estimated by the LIO at the time step $k$ be
\begin{align*}
    \bm\eta_k = ({}^{\text{B}}\mathbf{R}_{\text{O},k},{}^{\text B}\mathbf{t}_{\text{O},k},{}^{\text B}\mathbf{v}_{\text{O},k},{}^{\text B}\mathbf{b}^a_{\text{O},k},{}^{\text B}\mathbf{b}^g_{\text{O},k}),
\end{align*}
where ${}^{\text{B}}\mathbf{R}_{\text{O},k}$ and ${}^{\text B}\mathbf{t}_{\text{O},k}$ denote the same transformation variables as introduced in \eqref{eq:leg_fk}, 
${}^{\text B}\mathbf{v}_{\text{O},k} \in\mR^3$ is the linear velocity expressed in the world frame, and ${}^{\text B}\mathbf{b}^a_{\text{O},k}, {}^{\text B}\mathbf{b}^g_{\text{O},k}\in\mR^3$ are the accelerometer and gyroscope bias expressed in the world frame, respectively.

Given discrete IMU measurements $\big(\mathbf{a}_i,\, \boldsymbol{\omega}_i\big)$ collected between the $k$-th and $(k+1)$-th LiDAR scans, 
where $\mathbf{a}_i \in \mathbb{R}^3$ and $\boldsymbol{\omega}_i \in \mathbb{R}^3$ 
denote the specific acceleration and angular velocity measured in the IMU frame at time $t_i$ with sampling interval $\Delta t_i$, 
the pre-integrated relative motion takes the form
\begin{align*}
    \Delta {}^{\text{B}}\mathbf{R}_{\text{O},k} &= \prod_{i} \exp\!\left( \lfloor\boldsymbol{\omega}_i - {}^{\text B}\mathbf{b}^g_{\text{O},k}\rfloor_\times \, \Delta t_i \right), \\
    \Delta {}^{\text B}\mathbf{v}_{\text{O},k} &= \sum_{i} {}^{\text{B}}\mathbf{R}_{\text{O},k} \left( \mathbf{a}_i - {}^{\text B}\mathbf{b}^a_{\text{O},k} \right) \Delta t_i, \\
    \Delta {}^{\text B}\mathbf{t}_{\text{O},k} &= \sum_{i} {}^{\text B}\mathbf{v}_{\text{O},k} \, \Delta t_i 
        + \tfrac{1}{2} {}^{\text{B}}\mathbf{R}_{\text{O},k} \left( \mathbf{a}_i - {}^{\text B}\mathbf{b}^a_{\text{O},k} \right) \Delta t_i^2.
\end{align*}
Here, $\exp(\cdot)$ denotes the matrix exponential on the Lie algebra $\mathfrak{so}(3)$. 
The angular velocity vector $\boldsymbol{\omega}_i - \mathbf{b}^g_k \in \mathbb{R}^3$ is first mapped to a skew-symmetric matrix $\lfloor\boldsymbol{\omega}_i - \mathbf{b}^g_k\rfloor_\times \in \mathfrak{so}(3)$, 
and then the exponential map $\exp(\lfloor\boldsymbol{\omega}_i-  \mathbf{b}^g_k\rfloor_\times \Delta t_i)$ gives the corresponding incremental rotation in $\mathrm{SO}(3)$.

\subsection{The Manifold Constraints}
We assume the wheels contact firmly with the terrain. 
Consequently, the manifold constraint that regulates the pose estimate in vertical direction shall take the form
\begin{align}
    r_{\mathcal{M}}^{\text{L}}(\bm{\xi}_\text{L}):=\xi_{z,\text{L}}-r - f([\xi_{x,\text{L}},\xi_{y,\text{L}}]^\top;\hat{\mathbf{w}}^{RG})\approx \textbf{0},
    \label{eq:residual_manifold_left_wheel}
\end{align}
and the manifold constraint induced by the right wheel follows similarly.

Next, we construct the LIO framework and integrate the manifold constraints into the scan-matching optimization. 
Our LIO system structure follows the front-end of LIO-SAM \cite{liosam}. 
In particular, we use the IMU pre-integration module as mentioned in Sec.~\ref{sec:imu-preint} to provide motion priors between consecutive LiDAR frames. 
Moreover, we also use the factor graph in the IMU pre-integration module to update the linear acceleration and gyroscope bias.

In particular, the scan matching we use in this work follows a feature-based ICP formulation as that in LOAM \cite{loam} and LIO-SAM \cite{liosam}. To initialize the ICP optimization, we employ the IMU pre-integration between the two consecutive frames to estimate the relative pose. This relative pose is then added to the global pose of the previous frame, yielding an initial guess of the current frame's global transformation for the optimization solver.
More specifically, the ICP formulation minimizes 
\begin{align*}
    J_{\text{scan}}(\bm\eta_k) =&
       \sum_{\mathbf{p}^e_{s,k} \in \mathcal{F}^e_{k}} \| \mathbf{d}_{e} ({}^{\text{B}}\mathbf{R}_{\text{O},k},{}^{\text {B}}\mathbf{t}_{\text{O},k};\mathbf{p}^e_{s,k}) \|^2 \\
        &+
        \sum_{\mathbf{p}^p_{s,k} \in \mathcal{F}^p_{k}} \| \mathbf{d}_{p}({}^{\text{B}}\mathbf{R}_{\text{O},k},{}^{\text {B}}\mathbf{t}_{\text{O},k};\mathbf{p}^p_{s,k}) \|^2,
\end{align*}
with respect to ${}^{\text{B}}\mathbf{R}_{\text{O},k},{}^{\text B}\mathbf{t}_{\text{O},k}$,
where $\mathcal{F}^e_{k}$ and $\mathcal{F}^p_{k}$ denote the sets of edge and planar features in the LiDAR scan at the time step $k$.
$\mathbf{p}^e_{s,k}$ and $\mathbf{p}^p_{s,k}$ denote the $s$-th edge and planar feature points extracted from the scan at the time step $k$; $\mathbf{d}_{e} ({}^{\text{B}}\mathbf{R}_{\text{O},k},{}^{\text {B}}\mathbf{t}_{\text{O},k};\mathbf{p}^e_{s,k})$ and $\mathbf{d}_{p}({}^{\text{B}}\mathbf{R}_{\text{O},k},{}^{\text {B}}\mathbf{t}_{\text{O},k};\mathbf{p}^p_{s,k})$ are the corresponding point-to-line and point-to-plane residuals.

Now we add the manifold constraints \eqref{eq:residual_manifold_left_wheel} into the objective function as ``soft constraints" to regulate the pose estimation. In particular, the manifold ``soft constraints" with a penalty weight $\lambda_{\mathcal{M}}$ shall take the form
\begin{align*}
    J_{\mathcal{M}}(\bm\eta_k) = 
    \lambda_{\mathcal{M}} &\cdot (\left\| r_{\mathcal{M}}^L({}^{\text{B}}\mathbf{R}_{\text{O},k} \mathbf{h}_\text{L}(\mathbf{q}_k)+{}^{\text{B}}\mathbf{t}_{\text{O},k}) \right\|^2 \\
    &+
     \left\| r_{\mathcal{M}}^R({}^{\text{B}}\mathbf{R}_{\text{O},k} \mathbf{h}_\text{R}(\mathbf{q}_k)+{}^{\text{B}}\mathbf{t}_{\text{O},k}) \right\|^2).
\end{align*}
Thus, the total cost function for scan matching reads
\begin{equation}\label{eq:cost_fn}
    J_{\text{total}}(\bm\eta_k) =
    J_{\text{scan}}( \bm\eta_k) + J_{\mathcal{M}}(\bm\eta_k).
\end{equation}
Then we use Levenberg-Marquardt algorithm \cite{LM_algo} to minimize \eqref{eq:cost_fn} for scan matching. In particular, the gradient is needed to solve the problem. Since the gradients of point-to-line and point-to-plane residuals have been discussed in detail in \cite{liosam}, we will only discuss the new added $\frac{\partial}{\partial\bm\eta} r_{\mathcal{M}}(\bm\eta)$ here. 
In particular, we use the residual corresponds to the left wheel as example. By chain rule, the Jacobian of $J_{\mathcal{M}}(\bm\eta)$ with respect to the state $\bm\eta$ shall take the form $\frac{\partial J_{\mathcal{M}}}{\partial \bm\eta} = 2 \lambda_{\mathcal{M}} \cdot r_{\mathcal{M}} \cdot \frac{\partial r_{\mathcal{M}}}{\partial \bm\eta}$,
where
\begin{equation}
    \frac{\partial r_{\mathcal{M}}}{\partial \bm\eta} =
    \frac{\partial r_{\mathcal{M}}}{\partial \boldsymbol{\xi}_{\mathrm{L}}}
    \cdot
    \frac{\partial \boldsymbol{\xi}_{\mathrm{L}}}{\partial \bm\eta},\quad
    \frac{\partial r_{\mathcal{M}}}{\partial \boldsymbol{\xi}_{\mathrm{L}}} :=
    \begin{bmatrix}
        -\frac{\partial f}{\partial p_x}(\xi_{x,\text{L}}, \xi_{y,\text{L}}) \\
        -\frac{\partial f}{\partial p_y}(\xi_{x,\text{L}}, \xi_{y,\text{L}}) \\
        1
    \end{bmatrix},
    \label{eq:manifold_derivative}
\end{equation}
and
\begin{align}\label{eq:manidolf_grad_detail}
    \frac{\partial f}{\partial p_x}(\xi_{x,\mathrm{L}}, \xi_{y,\mathrm{L}}) 
    &= \sum_{j=1}^n [w_j \cdot 
       \frac{\xi_{x,\mathrm{L}} - x_j^c}{\sigma^2} \notag \\
    &\quad \times \exp\left(
       -\frac{\lVert (\xi_{x,\mathrm{L}}, \xi_{y,\mathrm{L}}) - \mathbf{x}_j \rVert^2}{2\sigma^2} 
       \right)], \nonumber \\
    \frac{\partial f}{\partial p_y}(\xi_{x,\mathrm{L}}, \xi_{y,\mathrm{L}}) 
    &= \sum_{j=1}^n [w_j \cdot 
       \frac{\xi_{y,\mathrm{L}} - y_j^c}{\sigma^2} \notag \\
    &\quad \times \exp\left(
       -\frac{\lVert (\xi_{x,\mathrm{L}}, \xi_{y,\mathrm{L}}) - \mathbf{x}_j \rVert^2}{2\sigma^2} 
       \right)].
\end{align}
Moreover, it holds that
\begin{align}\label{eq:manifold_grad_2}
    \frac{\partial \boldsymbol{\xi}_{\text{L}}}{\partial \bm\eta} &=
    \begin{bmatrix}
        -{}^{\text{B}}\mathbf{R}_\text{O} \cdot \lfloor{}^{\text B}\mathbf{t}_{\text{O}}\rfloor_\times & \mathbf{I}_{3 \times 3},
    \end{bmatrix}
\end{align}
where $\lfloor\cdot\rfloor_\times$ denotes the induced skew-symmetric matrix.

\section{GPU acceleration}
Our proposed LIO framework has a structure that can easily achieve a high parallel computation with GPU.
Consequently, it is able to execute efficiently with those computationally intensive components. In particular, two modules benefit from GPU implementation:
\begin{enumerate}
    \item \textbf{The RBF-related terms evaluation.} 
    The evaluation of the RBF surface involves computing exponential terms for each center–LiDAR point pairs. 
    This operation can be highly parallelized, therefore it is offloaded to the GPU to exploit data-level parallelism. 
    In our implementation,  the multiplication in height estimation \eqref{eq:fit_height} and entries in the RBF kernel matrix \eqref{eq:rbf_kernel} are calculated in parallel by GPU.
    In addition, the evaluation of the ``soft constraints" includes both the residuals \eqref{eq:residual_manifold_left_wheel} induced by the left and right legs and their gradients \eqref{eq:manifold_derivative}. In particular, recall that the term $f([\xi_{x,\text{L}},\xi_{y,\text{L}}]^\top;\hat{\mathbf{w}}^{RG})$ in \eqref{eq:residual_manifold_left_wheel} takes the form \eqref{eq:RBF_def}, where each component in the sum can be computed in parallel by GPU. Similarly, we can also use GPU to compute the components in the sum in \eqref{eq:manidolf_grad_detail} in parallel.

    \item \textbf{The Sparse kernel matrix inversion.} 
    The RBF kernel matrix is sparse due to the compact support of basis functions. 
    Thus, its inversion can be accelerated by using the \texttt{cuSOLVER} library from the CUDA Toolkit. 
    This library provides efficient sparse linear algebra solvers on GPU. In our implementation, the matrix inverse in \eqref{eq:kf_w_update}, \eqref{eq:active_rows_weights_update} and \eqref{eq:H_inverse_update} are all accelerated by GPU.

\end{enumerate}

\begin{figure}[!htpb]
  \centering
  \includegraphics[width=0.9\linewidth]{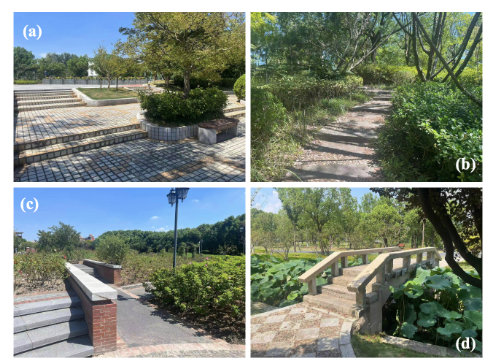}
  \caption{Dataset environments: (a) Staircase, (b) Artificial Hill, (c) Rose Garden, (d) Botanical Garden. These datasets include diverse terrain elevations, and each environment contains different stair structures to verify the $z$-axis localization performance of different method.}
  \label{fig:dataset-env}
\end{figure}

\section{Experiments}\label{sec:experiment}
We conduct the experiments on the Limx Tron1 legged-wheel robot in outdoor scenarios. Tron1, developed by Limx Dynamics, features a bipedal lower‐body drive with two selectable locomotion modes—\emph{normal} for flat terrain and \emph{off‐road} for stairs and uneven ground. In particular, the ``off‐road locomotion mode" is used in all of the experiments.
In addition, the following sensors are available for our experiment (as illustrated in Fig.~\ref{fig:robot-setting}).
\begin{itemize}
  \item \textbf{Mid360 LiDAR \& IMU:} mounted on a custom shelf attached to the robot's body, providing \SI{10}{\hertz} 3D LiDAR point clouds and \SI{200}{\hertz} inertial measurements.
  \item \textbf{GPS+RTK:} mounted upright atop the robot, delivering \SI{10}{\hertz} position fixes (latitude, longitude, altitude) of the ``ground-truth" trajectory.
  \item \textbf{Joint sensors:} joint angles are measured with a frequency of \SI{500}{\hertz} via the joint sensors.
\end{itemize}
Moreover, we test our method on a laptop equipped with an Intel i5-12490F processor, 32 GB of RAM, and an NVIDIA GeForce RTX 3070 GPU. 
\begin{figure}[!htpb]
    \centering
    \includegraphics[width=0.9\linewidth]{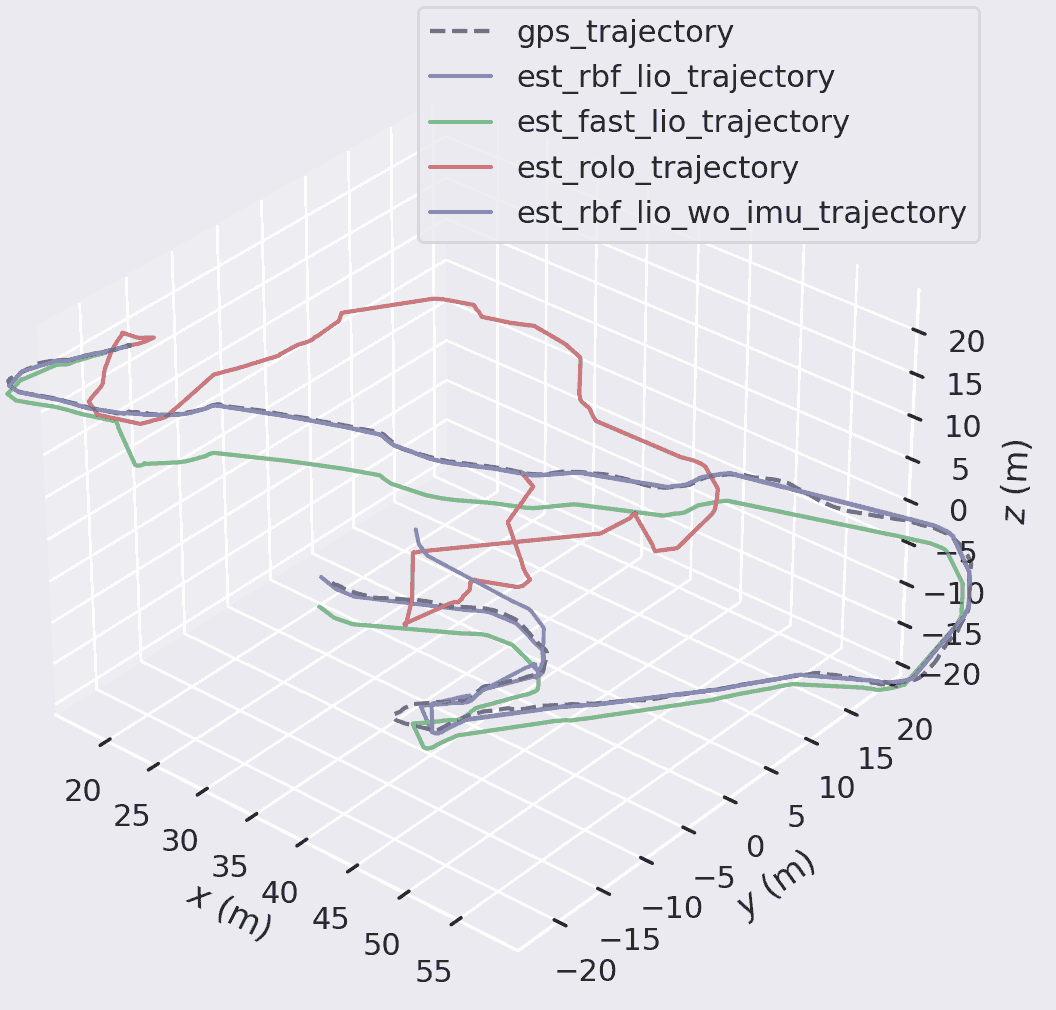}
    \caption{3D trajectories of different methods on the rose garden dataset. 
    While RBF-LIO and its IMU-free variant remain close to the GPS reference, 
    Fast-LIO2 and ROLO-SLAM exhibit noticeable deviations, especially around the staircase section.}
    \label{fig:traj}
\end{figure}

\begin{table*}[!t]
\centering
\caption{ATE\cite{ATE} (m) for RBF-LIO, Fast-LIO2\cite{fastlio2}, ROLO-SLAM\cite{roloslam}, and RBF-LIO (w.o. IMU)}
\label{tab:ape}
\begin{tabular}{| c | cc | cc | cc | cc |}
\hline
Sequence & \multicolumn{2}{c|}{RBF-LIO} & \multicolumn{2}{c|}{Fast-LIO2\cite{fastlio2}} & \multicolumn{2}{c|}{ROLO-SLAM\cite{roloslam}} & \multicolumn{2}{c|}{RBF-LIO (w.o. IMU)} \\
         & RMSE   & max      & RMSE      & max       & RMSE      & max      & RMSE      & max      \\
\hline
Staircase Scene 1   & 0.8018     & \textbf{1.2669}        & \textbf{0.7782}        & 1.2944        & 21.6712        & 29.2766        & 0.8320        & 1.6164        \\
Staircase Scene 2   & 0.7639     & 1.6539        & \textbf{0.7562}        & \textbf{1.6065}        & 36.7933        & 65.8538      & 34.1355        & 41.6277       \\
Artificial Hill     & \textbf{0.8472} & \textbf{1.6365} & 1.0027 & 1.7798 & 16.1785 & 24.1648 & 0.8860 & 1.6566 \\
Rose Garden              & \textbf{0.6273} & \textbf{1.0075} & 3.8791 & 7.1445 & 16.9995 & 27.4158 & 1.5601 & 8.7205 \\
Botanical Garden    &  1.0885   & 1.7083        & \textbf{0.9903}       & \textbf{1.5099}        & 22.3889       & 41.6459        & 1.8273        & 5.0241       \\
\hline
\end{tabular}

\vspace{2em}

\caption{RTE\cite{ATE} (m) for RBF-LIO, Fast-LIO2\cite{fastlio2}, ROLO-SLAM\cite{roloslam}, and RBF-LIO (w.o. IMU)}
\label{tab:rpe}
\begin{tabular}{| c | cc | cc | cc | cc |}
\hline
Sequence & \multicolumn{2}{c|}{RBF‐LIO} & \multicolumn{2}{c|}{Fast‐LIO2\cite{fastlio2}} & \multicolumn{2}{c|}{ROLO‐SLAM\cite{roloslam}} & \multicolumn{2}{c|}{RBF‐LIO (w.o. IMU)} \\
         & RMSE   & max      & RMSE      & max       & RMSE      & max      & RMSE      & max      \\
\hline
Staircase Scene 1   & \textbf{0.0464}     & \textbf{0.2128}        & 0.0464        & 0.2133        & 0.1694        & 1.1933        & 0.0790        & 0.8807        \\
Staircase Scene 2   & 0.0483     & 0.2337        & \textbf{0.0482}        & \textbf{0.2180}       & 0.2027        & 0.8776      & 0.0923        & 1.3317       \\
Artificial Hill     & \textbf{0.0600} & 0.4559 & 0.0608 & \textbf{0.4376} & 0.3977 & 8.7865 & 0.0675 & 0.5795 \\
Rose Garden              & \textbf{0.0371} & \textbf{0.1542} & 0.0477 & 0.4126 & 0.1881 & 0.6960 & 0.0645 & 1.3708 \\
Botanical Garden    &  \textbf{0.0566}   & \textbf{0.3589}        & 0.0567       & 0.3645        & 0.2104       & 0.8683        & 0.0608        & 0.9524       \\
\hline
\end{tabular}
\end{table*}

\subsection{The Dataset}
Due to the lack of open datasets for legged-wheel robots, we collect five real-world datasets that captures a wide variety of outdoor terrains, surface textures, and elevation profiles. The datasets also include abrupt elevation changes, mixed feature densities, and both structured and unstructured terrain. For each data sequence, we further include an RTK-corrected position coordinate (latitude, longitude, altitude) sequence measured by the GPS+RTK system. Hence it shall provide centimeter-level 3D ground-truth trajectories for accuracy evaluations. 
Moreover, we release the collected datasets as the source for legged-wheel robots' SLAM benchmark.
In particular, the content of the 5 datasets are as follows:
\begin{itemize}
  \item \textbf{Staircase Scene 1 \& 2}: comprise the routes that traverse flat paves and different staircases with step heights 5–15 cm, 1–8 steps, including single-pass and multiple back-and-forth traversals.
  \item \textbf{Artificial Hill}: comprises a route that traverses stone-slab paths with multiple single-step stairs, from the foot of the hill to the summit, and back down to the starting point from another path, exhibiting the height changes and varied terrain surface textures.
  \item \textbf{Rose Garden}: comprises a route that traverses short staircases, flat paves and grass slopes, with low lateral obstacles (flower beds, shrubs). In addition, the staircases are flanked by low walls, causing the sparse features around the staircases.
  \item \textbf{Botanical Garden}: comprises a long route in the botanical garden that traverses a suspension bridge, an arch bridge, multiple single-step stairs and a stone-slab path;
\end{itemize}
\textbf{The Ground Truth:}  
We use the GPS position measures fixed by RTK as the ground truth. Nevertheless, since the GPS+RTK system only delivers latitude, longitude, and altitude but no orientation, we will only compare the translation error between different algorithms.

\begin{figure}[!htbp] 
  \centering

  \includegraphics[width=0.95\linewidth]{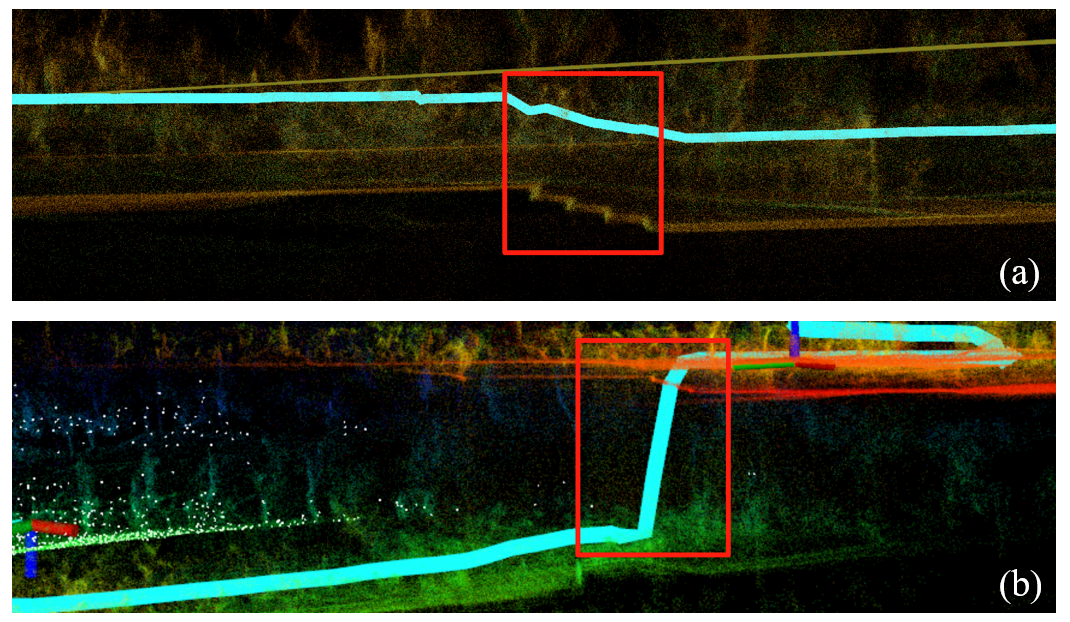}

  \caption{Visualization of the Rose Garden map: 
  (a) RBF mapping result (side view), 
  (b) Fast-LIO2 mapping result (side view).
  The staircase is highlighted by the red boxes, where Fast-LIO2 exhibits $z$-axis pose drift with layered artifacts, while RBF-LIO preserves the map consistency.}
  \label{fig:garden-results}
\end{figure}

\subsection{RBF-based Terrain Approximation Result}
\begin{figure}[!htbp]
    \centering
    \includegraphics[width=1\linewidth]{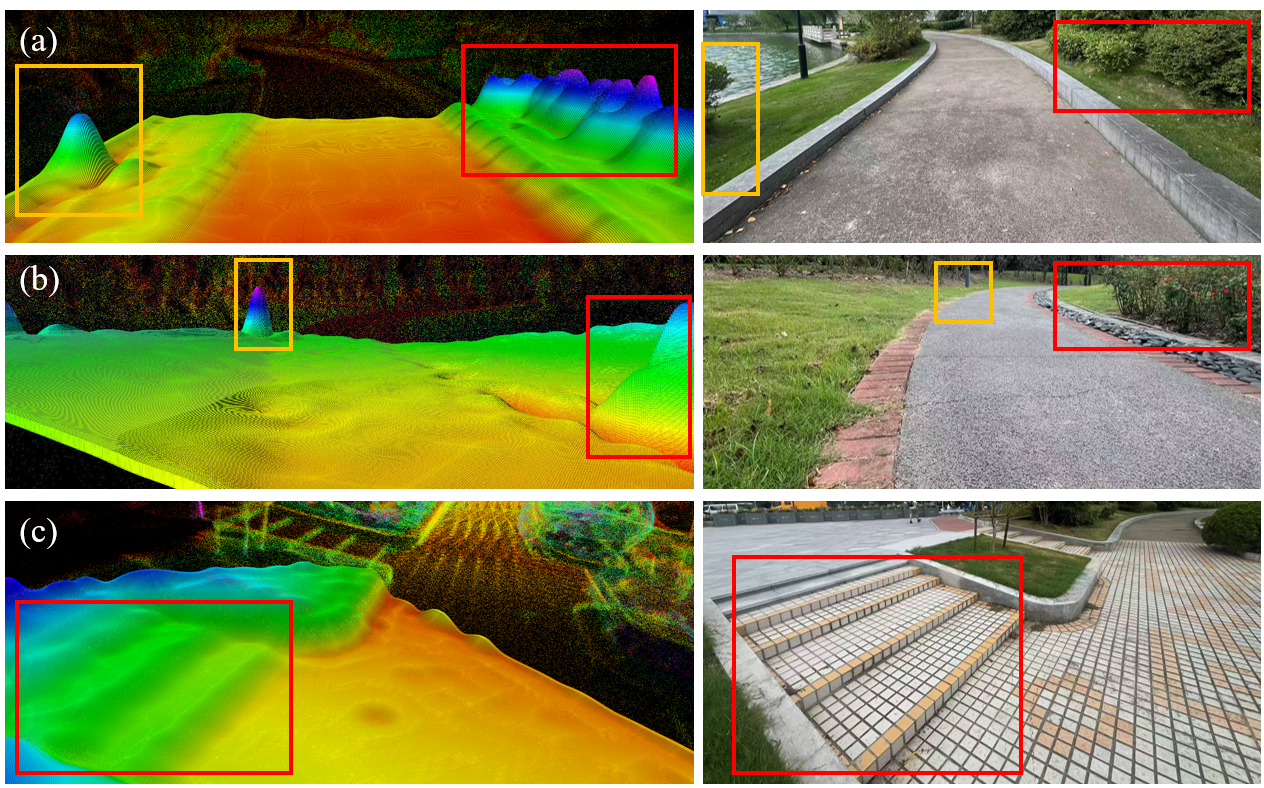}
    \caption{RBF-based terrain approximation in different environments: 
  (a) flat area, where the RBF model effectively captures large-scale planar surfaces; 
  (b) garden pathway, where the left slope is reconstructed, a utility pole on the slope is captured as a steep hump, and the edge between the pathway and the grass on the right is clearly captured; 
  (c) staircase scene, where multiple discrete elevation changes corresponding to stair steps are distinctly captured.
  The same objects are marked by boxes with the same color.
  }
    \label{fig:height}
\end{figure}

\begin{figure}[!htbp]
    \centering
    \includegraphics[width=1\linewidth]{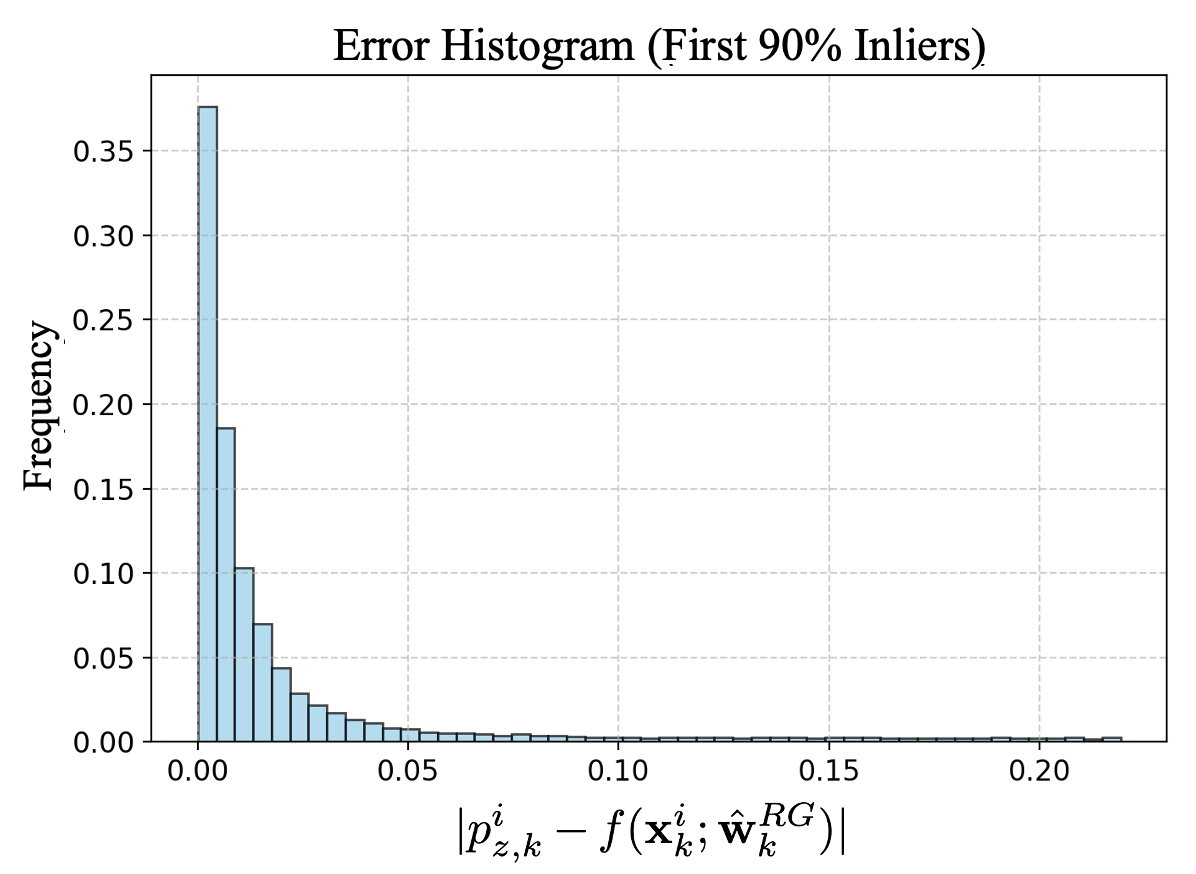}
    \caption{The errors histogram of the terrain approximation on the Rose Garden dataset after excluding the top 10\% outliers.}
    \label{fig:error_height}
\end{figure}

In our experiments, we set the resolution of the RBF center mesh to 0.07 m , the bandwidth parameter $\sigma$ to 0.04 m, and $\sigma_\epsilon$ to 0.1.
We first qualitatively demonstrate the result of RBF approximation on 3 different terrains in Fig.~\ref{fig:height}.
In particular, for the flat area (Fig.~\ref{fig:height}a), our method can approximate the broad planar surfaces, the bushes and the flower bed well.
For the garden pathway (Fig.~\ref{fig:height}b), the approximated terrain from our method captures the small bumps on the left slope, the utility pole on the slope, and the boundaries between the pathway and the adjacent grass field.
Moreover, in the staircase scenario (Fig.~\ref{fig:height}c), the method distinctly reconstructs multiple discrete elevation changes, corresponding to individual stair steps.
To further quantify the approximation accuracy, we report the error histogram of the RBF-based terrain approximation on the dataset Rose Garden.
In particular, the RBF approximation provides a continuous, dense presentation of the terrain, yet we can only evaluate the error on discrete points. To this end, we take the LiDAR point cloud $\mathbf{p}^i_k=[\mathbf{x}_k^{i\top},p_{z,k}^i]^\top$ expressed in the world frame as the ground truth and let $|p_{z,k}^i-f(\mathbf{x}_k^{i};\hat{\mathbf{w}}_k^{RG})|$ as the approximation error.
In addition, since there are some LiDAR point cloud outliers due to water puddle reflection, we show the error histogram after excluding the top 10\% outliers in Fig.~\ref{fig:error_height}.
Consequently, 92.88\% of the errors fall below 0.05 m. It indicates that the RBF-based terrain approximation performs well when the robot traverse the staircase and grass. It also coincide with the observation in Fig.~\ref{fig:garden-results} that RBF-LIO successfully mitigates localization drift when traversing the staircase, where Fast-LIO2 have an obvious localization drift on $z$-axis.

\begin{figure}[!htbp]
    \centering
    \includegraphics[width=0.9\linewidth]{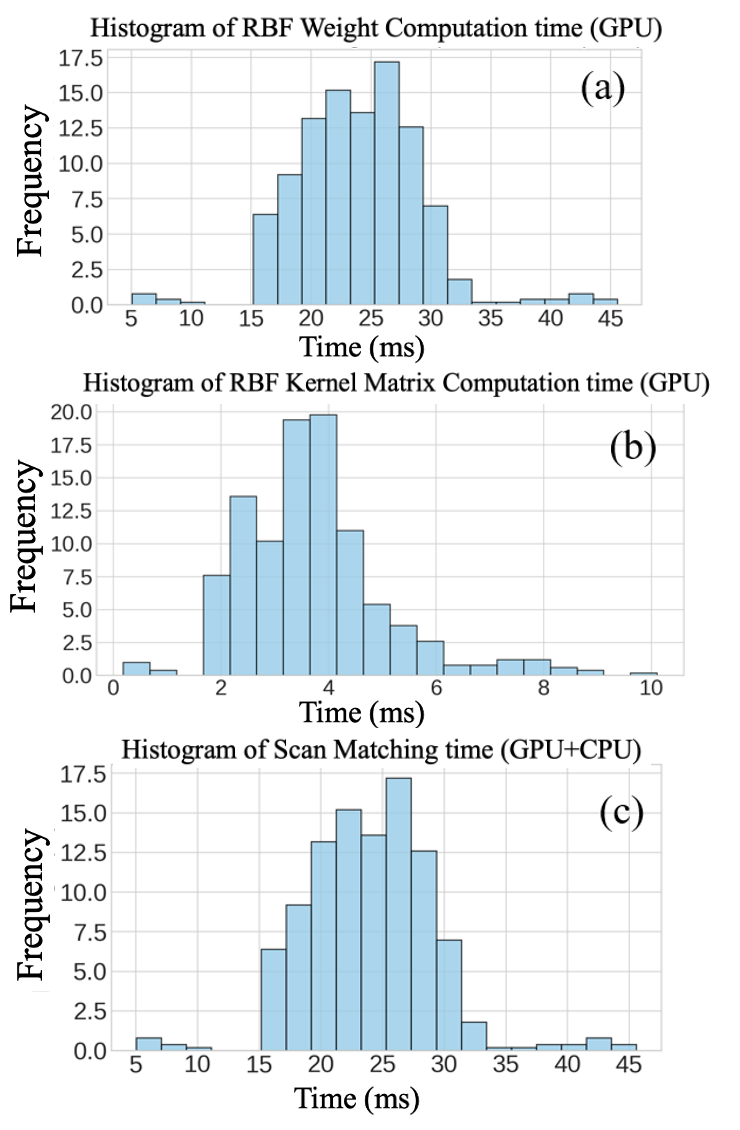}
    \caption{Time consumption histogram of the modules in our method.}
    \label{fig:hist}
\end{figure}

Moreover, to demonstrate the real-time performance of our method, we present the time consumption histogram of the modules in Fig.~\ref{fig:hist}.
Consequently, the overall framework operates at a frequency of 10 Hz and thus it meets the real-time requirements.

\begin{table*}[!t]
\centering
\caption{ATE\cite{ATE} (m) on $z$-axis for RBF-LIO, Fast-LIO2\cite{fastlio2}, ROLO-SLAM\cite{roloslam}, and RBF-LIO (w.o. IMU)}
\label{tab:ape_z}
\begin{tabular}{| c | cc | cc | cc | cc |}
\hline
Sequence & \multicolumn{2}{c|}{RBF‐LIO} & \multicolumn{2}{c|}{Fast‐LIO2\cite{fastlio2}} & \multicolumn{2}{c|}{ROLO‐SLAM\cite{roloslam}} & \multicolumn{2}{c|}{RBF‐LIO (w.o. IMU)} \\
         & RMSE   & max      & RMSE      & max       & RMSE      & max      & RMSE      & max      \\
\hline
Staircase Scene 1      & \textbf{0.0644}        & \textbf{0.3443}    & 0.1149     & 0.4307    & 1.0336        & 2.4170        & 0.1149        & 0.4246        \\
Staircase Scene 2   & \textbf{0.0657}     & \textbf{0.3725}        & 0.0764        & 0.3982        & 3.8083        & 9.8349      & 0.7983        & 1.4762       \\
Artificial Hill     & \textbf{0.1949} & \textbf{0.4835} & 0.2078 & 0.4678 & 6.4400 & 11.2143 & 0.2636 & 0.5856 \\
Rose Garden              & \textbf{0.1162} & \textbf{0.2990} & 3.5239 & 7.0845 & 1.5915 & 4.8372 & 0.1515 & 0.6599 \\
Botanical Garden    & \textbf{0.2188 }      & \textbf{0.4851}  &  0.3017   & 0.6276                & 3.3272       & 7.2956        & 0.3035        & 0.6279       \\
\hline
\end{tabular}

\vspace{2em}

\caption{RTE\cite{ATE} (m) on $z$-axis for RBF-LIO, Fast-LIO2\cite{fastlio2}, ROLO-SLAM\cite{roloslam}, and RBF-LIO (w.o. IMU)}
\label{tab:rpe_z}
\begin{tabular}{| c | cc | cc | cc | cc |}
\hline
Sequence & \multicolumn{2}{c|}{RBF‐LIO} & \multicolumn{2}{c|}{Fast‐LIO2\cite{fastlio2}} & \multicolumn{2}{c|}{ROLO‐SLAM\cite{roloslam}} & \multicolumn{2}{c|}{RBF‐LIO (w.o. IMU)} \\
         & RMSE   & max      & RMSE      & max       & RMSE      & max      & RMSE      & max      \\
\hline
Staircase Scene 1   & \textbf{0.0135}     & 0.1152        & 0.0141        & \textbf{0.1087}        & 0.0401        & 0.3205        & 0.0140        & 0.1146        \\
Staircase Scene 2   & \textbf{0.0111}     & \textbf{0.0881}        & 0.0115        & 0.0996       & 0.0650        & 0.3859      & 0.0159        & 0.1782       \\
Artificial Hill     & \textbf{0.0247}   & 0.1616   & 0.0252   & \textbf{0.1544} & 0.0993   & 0.7038   & 0.0251   & 0.1569   \\
Rose Garden              & \textbf{0.0098} & \textbf{0.0830} & 0.0307 & 0.4078 & 0.0530 & 0.2968 & 0.0104 & 0.1164 \\
Botanical Garden    &  \textbf{0.0180}   & \textbf{0.3500}        & 0.0182       & 0.3563        & 0.0639       & 0.3172        & 0.0181        & 0.3479       \\
\hline
\end{tabular}
\end{table*}

\subsection{Comparison Results}
Our proposed RBF-LIO is compared with two representative baselines. 
First, we compare our method with Fast-LIO2 \cite{fastlio2}, the state-of-the-art tightly-coupled LiDAR–inertial odometry framework. 
In particular, Fast-LIO2 has shown the outstanding ability of localization and mapping in the vehicles and drones that have smooth trajectories. 
To guarantee a fair comparison, at the beginning of each experiment, the robot remains stationary for a period of time to let Fast-LIO2 complete its state initialization.
In addition, we also compare our method with ROLO-SLAM\cite{roloslam}, a LiDAR-only approach specifically designed to mitigate $z$-axis pose drift. And it does not incorporate the IMU measurements. To guarantee a fair comparison, we compare it with an IMU-ablated variant of our RBF-LIO (denoted as RBF-LIO w.o. IMU). 

Moreover, we keep the following variables the same in all the comparisons:
\begin{itemize}
  \item \textbf{The sensor measurements:} All methods use the same Mid360 LiDAR scans (10 Hz). RBF-LIO and Fast-LIO2 both use the IMU measurements (200 Hz). In addition, RBF-LIO also use the joint sensor measurements (500 Hz).
  \item \textbf{System parameters:} All the parameters, including the LiDAR point cloud down-sample rate, the measurement noise covariance, etc, are set identical in all methods.
\end{itemize}

To quantitatively compare the performance, we compare the methods' Absolute Trajectory Error (ATE) and Relative Trajectory Error (RTE) that are introduced in \cite{ATE}. In particular, we illustrate the overall ATE and RTE of the methods in Table.~\ref{tab:ape} and \ref{tab:rpe}. 
In addition, to show the effectiveness of the ``soft constraints", we also illustrate the ATE and RTE of the $z$ component of the trajectories produced by different methods in Table.~\ref{tab:ape_z} and \ref{tab:rpe_z}.
The best results are highlighted with bold font.

In scenarios that have rich environment features when encounter abrupt height changes (Botanical Garden and Staircase Scene 1 \& 2), RBF-LIO's ATE is comparable with Fast-LIO2. It indicates that Fast-LIO2 can accurately estimate the robot's location in the rich-feature environments. Also, in scenarios that have gentle slopes (Botanical Garden), RBF-LIO and Fast-LIO2 have a similar performance. It further shows that the additional ``soft constraints" derived from the contact points approximation (see Fig~\ref{fig:wheel}) does not compromise the accuracy of the LIO. However, in the scenarios that have continuous height changes (Artificial Hill) or sparse features when abrupt height changes occur (Rose Garden), RBF-LIO reduces both ATE and RTE significantly. The improvements are most pronounced along the vertical dimension: while Fast-LIO2 suffer from progressive $z$-axis pose drift (see Table.~\ref{tab:ape_z} and \ref{tab:rpe_z}) and map layering artifacts (see Fig.~\ref{fig:garden-results}(b)), RBF-LIO continuously corrects height estimates through the ``soft constraints", thereby preserving a global consistent map (see Fig.~\ref{fig:garden-results}(a)). On the other hand, our method also outperforms ROLO-SLAM, which explicitly optimizes for vertical errors.
Consequently, the above comparison demonstrates the fact that our method not only outperforms the existing LIO, but also exceeds the performance of dedicated approaches for reducing the vertical localization errors. 

\begin{figure}[!htpb]
    \centering
    \includegraphics[width=\linewidth]{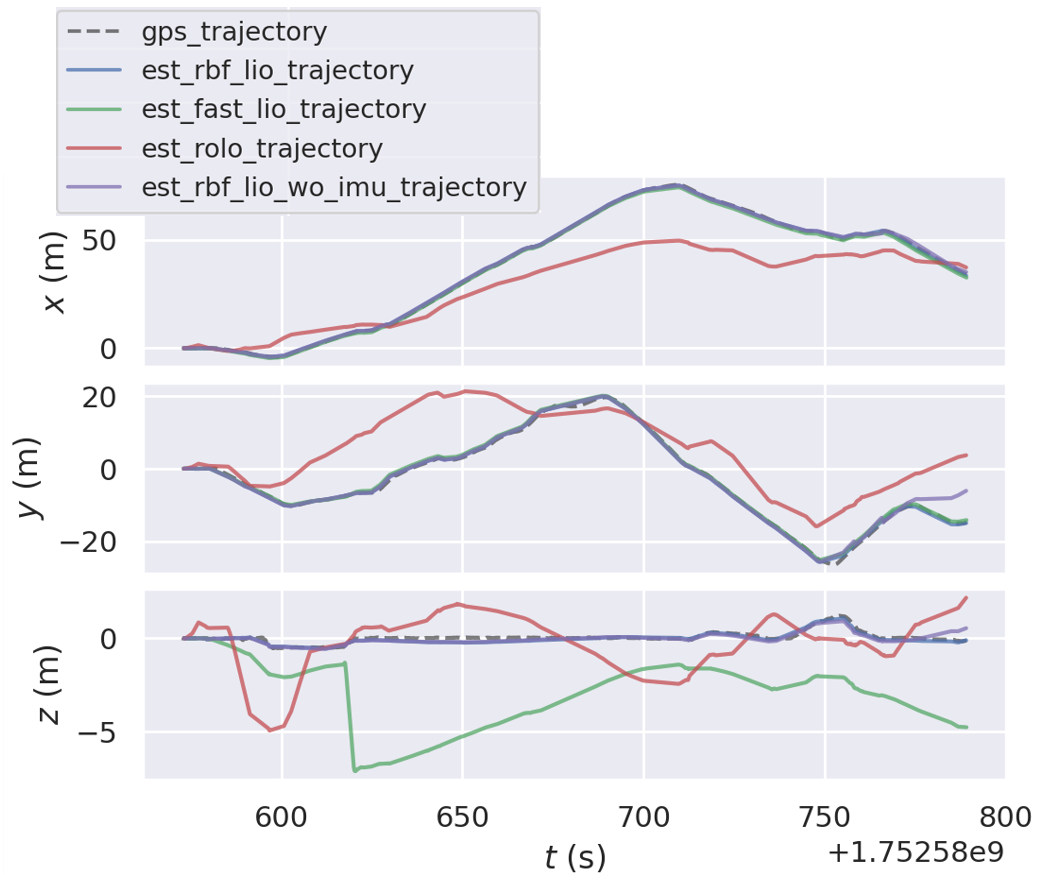}
    \caption{The $x$, $y$, and $z$ components of the trajectories generated by different methods in the Rose Garden scenario of the dataset.
    Fast-LIO2 and ROLO-SLAM suffer from significant drift in the vertical ($z$) direction, 
    particularly when traversing staircases, whereas RBF-LIO maintains an accurate position estimate.
    }
    \label{fig:plot}
\end{figure}

\section{Conclusion And Future Work}
In this paper, we propose RBF-LIO, a LiDAR–inertial odometry framework that incorporates terrain constraints to improve the vertical localization accuracy. In particular, our proposed method approximates the terrain with RBFs based on moment conditions, and the problem is further rewritten into a recursive ridge regression problem. Consequently, we get a manifold that approximates the terrain and further introduces the manifold as ``soft constraints" into the gradient-based LIO optimization to mitigate the localization error on the $z$-axis.
Such method significantly suppresses $z$-axis pose drift in the environments with large height differences and rapid elevation changes, compared to the state-of-the-art Fast-LIO2 and ROLO-SLAM. Consequently, the added ``soft constraints" enables continuous height corrections for the LIO, yielding coherent, high-fidelity maps even under abrupt stair ascents and feature-scarce conditions.
Moreover, we release a dataset that contains five scenarios for further legged-wheel robot's SLAM research.

The limitations of our method is as follows. The current RBF formulation for the terrain is a height function of two‐dimensional coordinates, which is not suitable for environments with multi‐layer structures (e.g., overpasses or stacked floors). In future work, we will explore more expressive surface representations to handle such complex geometries. Moreover, we intend to integrate continuous approximation functions like RBF into reinforcement learning frameworks for legged–wheel robots locomotion, enabling end‐to‐end training of perceptive control policies that leverage learned terrain priors.


\bibliographystyle{Bibliography/IEEEtranTIE}
\bibliography{refs}
	
\begin{IEEEbiography}[{\includegraphics[width=1in,height=1.25in,clip,keepaspectratio]{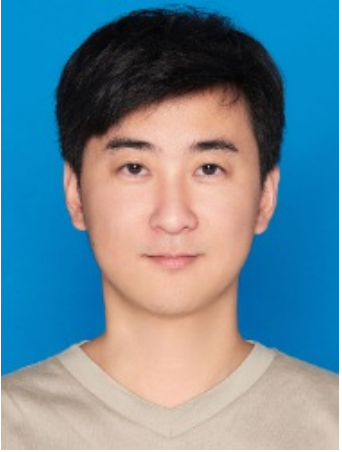}}]
{Yizhe Liu}
was born in Kunming, Yunnan, China, in December 2000. He received the B.S. degree in automation in June 2023 from Shanghai Jiao Tong University, Shanghai, China, where he is currently pursuing the M.S. degree in control engineering. His major field of study is Simultaneous Localization And Mapping (SLAM) and legged–wheel robot control.
\end{IEEEbiography}

\begin{IEEEbiography}[{\includegraphics[width=1in,height=1.25in,clip,keepaspectratio]{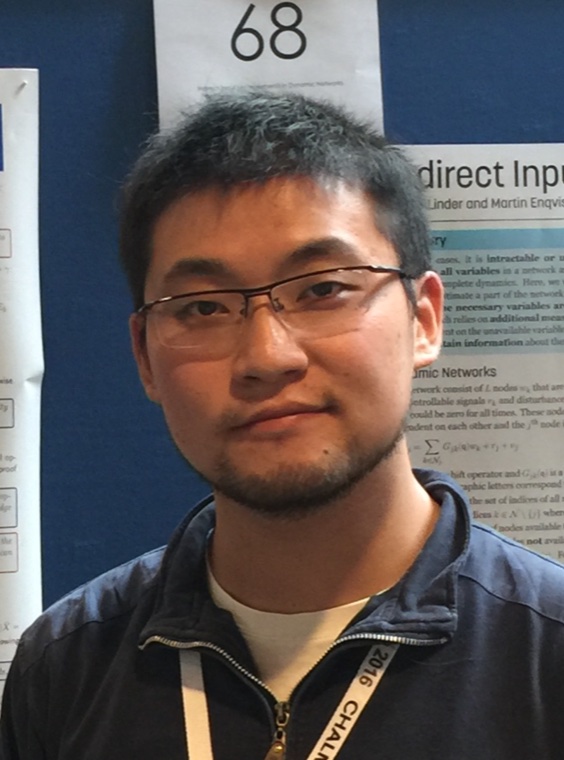}}]
{Han Zhang} was born in Shanghai, China. He received both his B.S. and M.S. degree from the Dept. of Automation in Shanghai Jiao Tong University in 2011 and 2014 respectively. He got his Ph.D. in Applied and Computational Mathematics specialized in Optimization and Systems Theory from Dept. of Mathematics, KTH Royal Institute of Technology, Sweden in 2019.

After spending two years in industry, he is now a tenure track associate professor in School of Automation and Intelligent Sensing, Shanghai Jiao Tong University. He is currently with the research group of Autonomous Robot Lab.

His main research interest includes but not limited to inverse optimal control, optimal control, SLAM with applications to autonomous vehicles, legged robots, rehabilitation and assistive robots.
\end{IEEEbiography}

\end{document}